\documentclass{article}

\usepackage{PRIMEarxiv}

\usepackage[utf8]{inputenc} 
\usepackage[T1]{fontenc}    
\usepackage{hyperref}       
\usepackage{url}            
\usepackage{booktabs}       
\usepackage{amsfonts}       
\usepackage{nicefrac}       
\usepackage{microtype}      
\usepackage{lipsum}
\usepackage{placeins}
\usepackage[numbers]{natbib}
\usepackage{fancyhdr}       
\usepackage{graphicx}       
\graphicspath{{media/}}     

\title{Comparative Efficiency Analysis of Lightweight Transformer Models: 
A Multi-Domain Empirical Benchmark for Enterprise NLP Deployment}

\author{
  Muhammad Shahmeer Khan \\
  School of Computing and Mathematics \\
  Ulster University \\
  Manchester, United Kingdom \\
  khan-MS15@ulster.ac.uk
}

\begin{document}
\maketitle

\begin{abstract}
In the rapidly evolving landscape of enterprise natural language processing (NLP), the demand for efficient, lightweight models capable of handling multi-domain text automation tasks has intensified. This study conducts a comparative analysis of three prominent lightweight Transformer models—DistilBERT, MiniLM, and ALBERT—across three distinct domains: customer sentiment classification, news topic classification, and toxicity/hate speech detection. Utilizing datasets from IMDB, AG News, and the Measuring Hate Speech corpus, we evaluated performance through accuracy-based metrics (accuracy, precision, recall, F1-score) and efficiency metrics (model size, inference time, throughput, and memory usage). Key findings reveal that no single model dominates all performance dimensions: ALBERT achieves the highest task-specific accuracy in multiple domains, MiniLM excels in inference speed and throughput, and DistilBERT demonstrates the most consistent accuracy across tasks while maintaining competitive efficiency. All results reflect controlled fine-tuning under fixed enterprise-oriented constraints rather than exhaustive hyperparameter optimization.These results underscore trade-offs in accuracy versus efficiency, recommending MiniLM for latency-sensitive enterprise applications, DistilBERT for balanced performance, and ALBERT for resource-constrained environments.
\end{abstract}


\section{Introduction}
Enterprises increasingly rely on NLP for automating text-based processes, such as sentiment analysis in customer feedback, topic categorization in news feeds, and content moderation for online safety. However, deploying large Transformer models like BERT in resource-limited settings poses challenges in terms of computational cost, latency, and memory usage,which can hinder scalability in enterprise environments \cite{sanh2019distilbert,bhattacharjee2023transformers}, where real-time processing and cost efficiency are paramount.Lightweight models address these issues by reducing parameters and inference times while preserving much of the performance, making them ideal for deployment on edge devices or in cloud-constrained setups.
While prior work extensively benchmarks large models like BERT and RoBERTa, comparative evaluation of lightweight Transformer models across multiple enterprise-relevant domains remains underexplored. Existing studies typically benchmark on a single dataset or task, limiting real-world applicability. This creates a gap in research focused on lightweight alternatives optimized for multi-domain enterprise use cases, where models must generalize across varied text types while maintaining efficiency. \\
Our contributions include: \\
(1) a systematic comparison of DistilBERT, MiniLM, and ALBERT on three enterprise-relevant domains \\
(2) integration of accuracy and efficiency metrics to highlight practical trade-offs and \\
(3) recommendations for enterprise deployment based on empirical results.

\
This work does not propose new model architectures or training algorithms. Instead, its primary contribution lies in a deployment-oriented empirical benchmark of widely adopted lightweight Transformer models under consistent enterprise constraints. By evaluating accuracy and efficiency trade-offs across multiple text classification domains using identical fine-tuning protocols, this study provides practical decision guidance for model selection in real-world enterprise NLP systems, where latency, throughput, and resource efficiency are often prioritized over marginal accuracy improvements.

\section{Related Work}

\subsection{Transformer Models in Enterprise NLP}
Transformer architectures have become foundational in NLP \cite{vaswani2017attention}, powering applications ranging from semantic understanding to document classification. However, large models such as BERT and RoBERTa impose substantial computational overhead,limiting their suitability for latency-sensitive and resource-constrained enterprise environments \cite{bhattacharjee2023transformers}.
 \\
This has driven demand for lightweight, deployable variants that maintain strong performance while reducing inference latency, memory usage, and operational costs. \\
In enterprise deployments — such as CRM analytics, risk assessment systems, and content moderation pipelines — throughput and cost per prediction often matter more than marginal accuracy improvements. This trend motivates the development and evaluation of parameter-efficient transformer variants optimized for real-time production workloads.

\subsection{Lightweight Transformer Architectures}
Several architectures have been proposed to address the computational limitations of full-scale models: \\ \\ 
\textbf{DistilBERT} \cite{sanh2019distilbert} reduces BERT’s size by 40 percent through knowledge distillation, retaining approximately 97 percent of performance while halving inference time. Its balance of accuracy and speed makes it one of the most widely adopted compact models in real-world NLP systems. \\ \\
\textbf{MiniLM} (Wang et al., 2020) introduces deep self-attention distillation, compressing attention relationships rather than full hidden states. This method enables significantly smaller models with competitive accuracy across benchmarks, particularly excelling in inference speed due to reduced hidden dimensions. \\ \\
\textbf{ALBERT} (Lan et al., 2020) achieves parameter efficiency through factorized embeddings and cross-layer parameter sharing, reducing model size dramatically (down to 12M parameters for ALBERT-Base). While beneficial for memory footprint, shared parameters introduce sequential dependencies that may increase inference latency. \\ \\
Collectively, these architectures represent the current landscape of lightweight transformers tailored for scalable deployment.

\subsection{Multi-Domain NLP Evaluation}

Benchmark studies typically evaluate models on a narrow set of tasks or a single dataset, such as sentiment analysis or question answering. While useful academically, this approach does not reflect the multi-domain variability seen in enterprise environments.
Recent works emphasize the importance of evaluating models across diverse classification tasks \cite{zhang2024lightweight,patel2024deltran}
 to test generalization capabilities. However, these studies either: 
\begin{itemize}
    \item do not incorporate efficiency metrics (e.g inference speed, throughput),
    \item focus on a single domain,
    \item or benchmark larger transformer architectures rather than lightweight ones.
\end{itemize}
Thus, a comprehensive cross-domain evaluation of lightweight transformer models — under enterprise-relevant constraints — remains largely unexplored.

\section{Methodology}

This section describes the lightweight Transformer models, datasets, experimental setup, and evaluation protocol used in our study.

\subsection{Models}

We evaluated three widely used lightweight Transformer architectures that are practical candidates for enterprise deployment:

\begin{itemize}
    \item \textbf{DistilBERT (\texttt{distilbert-base-uncased})}:  
    A distilled version of BERT with approximately 66 million parameters and six encoder layers. It is trained via knowledge distillation from BERT, retaining most of its language understanding capacity while reducing model depth and computational cost.
    
    \item \textbf{MiniLM (\texttt{microsoft/MiniLM-L12-H384-uncased})}:  
    A compact Transformer model that employs deep self-attention distillation to compress larger models into a smaller network (approximately 33 million parameters). This approach focuses on preserving attention patterns while reducing hidden dimensionality, enabling efficient inference.
    
    \item \textbf{ALBERT (\texttt{albert-base-v2})}:  
    A parameter-efficient variant of BERT with approximately 12 million parameters. ALBERT uses factorized embeddings and extensive cross-layer parameter sharing, substantially reducing the number of trainable parameters while maintaining model depth.
\end{itemize}

For each architecture,we used publicly available pre-trained checkpoints from the Hugging Face Transformers library and fine-tune them on each downstream task using a linear classification head.

\subsection{Datasets and Tasks}

We considered three enterprise-relevant domains, each formulated as a supervised text classification task:

\begin{itemize}
    \item \textbf{Domain 1 – Customer Sentiment Classification (IMDB)}:  
    The IMDB movie reviews dataset consisting of 50,000 reviews labeled as either positive or negative was used. The standard dataset split is used, with 25,000 examples for training and 25,000 for testing.
    
    \item \textbf{Domain 2 – News Topic Classification (AG News)}:  
    The AG News corpus contains 120,000 training samples and 7,600 test samples labeled into four categories: World, Sports, Business, and Sci/Tech. This task approximates enterprise content routing and topic-based automation pipelines.
    
    \item \textbf{Domain 3 – Toxicity / Hate Speech Detection (Measuring Hate Speech)}:  
     The Measuring Hate Speech corpus \cite{kennedy2022measuring}, which provides a continuous \texttt{hate\_speech\_score} for each comment was used to align with practical content moderation taxonomies,we discretized this score into three classes:
\end{itemize}

\begin{enumerate}
    \item \textbf{Neutral}: score $< 0.3$
    \item \textbf{Offensive}: $0.3 \leq$ score $< 0.7$
    \item \textbf{Hate Speech}: score $\geq 0.7$
\end{enumerate}

Since the Measuring Hate Speech dataset is released with a single training split, we created an 80/20 stratified split into training and validation subsets using a fixed random seed of 42. The validation subset is used as the evaluation set for this domain.

Across all domains, tasks are treated as standard supervised text classification problems with a single label per instance.

\subsection{Experimental Setup}

All experiments are implemented in Python using the Hugging Face Transformers and Datasets libraries. Training and evaluation are performed on a single NVIDIA T4 GPU to ensure consistent runtime conditions using Google Colab. Accordingly, our findings should be interpreted as model behavior under rapid fine-tuning constraints rather than fully converged performance.

For each model–dataset combination, we used the following shared hyperparameters:

\begin{itemize}
    \item Optimizer: AdamW (default settings from the Transformers \texttt{Trainer})
    \item Learning rate: $2 \times 10^{-5}$
    \item Number of epochs: 3
    \item Per-device training batch size: 16
    \item Per-device evaluation batch size: 16
    \item Weight decay: 0.01
    \item Logging frequency: every 200–500 steps (dataset-dependent)
\end{itemize}

Tokenization is performed using each model’s respective tokenizer. For all datasets, the input text field is \texttt{text}. We enable truncation with a fixed maximum sequence length of 128 tokens, which covers the majority of text lengths across all datasets while significantly improving inference efficiency in deployment-oriented settings. Padding is applied dynamically at batch time using a \texttt{DataCollatorWithPadding} to minimize unnecessary computation on shorter sequences. All models are fine-tuned for three epochs to ensure stable convergence across architectures while maintaining practical training times suitable for enterprise deployment scenarios.

For the IMDB and AG News datasets, we used the official test splits for evaluation. For the Measuring Hate Speech dataset, the held-out 20\% validation subset serves as the evaluation set.

All models are fine-tuned independently under identical hyperparameters to ensure a fair comparison of accuracy and efficiency.

\subsection{Evaluation Protocol}

We evaluate each model using two categories of metrics.

\paragraph{Accuracy-based metrics.}
To assess classification quality, we report:
\begin{itemize}
    \item Accuracy, defined as the proportion of correctly classified instances.
    \item Weighted Precision, Recall, and F1-score, where per-class metrics are weighted by class support. This is particularly important for the hate speech detection task, which exhibits class imbalance.
    \item Confusion matrices computed separately for each model and domain to analyze misclassification patterns. Due to space constraints, these matrices are provided in the supplementary materials (Appendices A–C).
\end{itemize}

\paragraph{Efficiency metrics.}
To evaluate deployment feasibility under enterprise constraints, we measured:
\begin{itemize}
    \item Model size (MB), computed as the number of trainable parameters multiplied by 4 bytes (FP32 representation).
    \item Inference time per sample (milliseconds), measured by timing predictions over a fixed subset of 200 evaluation examples and averaging wall-clock time per instance.
    \item Throughput (samples per second), computed as the number of evaluated samples divided by total inference time.
    \item RAM usage (MB), measured as the resident memory of the Python process after evaluation.
\end{itemize}

\paragraph{Efficiency measurement details.}
All inference measurements were conducted with gradient computation disabled using \texttt{torch.no\_grad()}.
Inference latency was measured using batch size 1 to reflect real-time deployment scenarios, following an initial warm-up phase to mitigate kernel compilation and caching effects.

Throughput measurements reflect aggregate processing rates under identical runtime conditions.
RAM usage is reported as approximate process-level memory consumption and includes framework overhead; it should therefore be interpreted as a coarse indicator rather than a model-isolated metric.
All efficiency measurements are hardware-dependent and should be interpreted as relative comparisons rather than absolute deployment guarantees. Results are reported under identical runtime conditions to ensure fairness across architectures, and conclusions are drawn based on consistent efficiency rankings rather than raw latency values. FP32 inference is used throughout to maximize reproducibility and comparability across models without introducing precision-specific optimizations.

\paragraph{Reproducibility and reporting.}
All dataset splits and training runs use a fixed random seed of 42 to ensure reproducibility. Because the primary objective of this study is deployment-oriented benchmarking, we report single-run results without repeated trials or statistical significance testing. In enterprise environments, throughput, latency, and resource efficiency are often prioritized over marginal performance variance, and the deterministic experimental setup ensures that relative comparisons remain practically meaningful.

To improve interpretability, results for each domain are summarized using a single consolidated table combining accuracy-based and efficiency metrics, accompanied by a bar chart visualizing key efficiency indicators.Rather than emphasizing marginal accuracy differences, our analysis focuses on consistent relative rankings and directional performance trends across models and domains under identical training conditions.We therefore do not claim statistical superiority for closely spaced accuracy values, but instead analyze stability and trade-offs relevant to deployment-oriented decision making.
This presentation facilitates direct comparison and supports informed model selection for real-world enterprise NLP deployments.

\section{Results}

\subsection{Domain 1: Customer Sentiment Classification (IMDB)}

In the sentiment classification task, ALBERT achieves the highest accuracy and F1-score, indicating that parameter-efficient architectures can perform competitively on binary sentiment tasks when adequately fine-tuned. DistilBERT demonstrates stable performance, while MiniLM exhibits slightly lower accuracy but maintains lower inference latency and higher throughput. ALBERT’s compact footprint is particularly advantageous for memory-constrained enterprise deployments, whereas MiniLM offers a favorable balance between efficiency and predictive performance. 
\begin{table}[t]
\centering
\caption{Performance and Efficiency Metrics — Domain 1 (IMDB Sentiment)}
\label{tab:domain1}
\resizebox{\columnwidth}{!}{
\begin{tabular}{lccccccc}
\toprule
\textbf{Model} 
& \multicolumn{4}{c}{\textbf{Accuracy Metrics}} 
& \multicolumn{3}{c}{\textbf{Efficiency Metrics}} \\
\cmidrule(lr){2-5} \cmidrule(lr){6-8}
& Acc. & Prec. & Rec. & F1 
& Size (MB) & Inf. (ms) & Thpt. \\
\midrule
DistilBERT 
& 0.932 & 0.929 & 0.936 & 0.932
& 255.42 & 14.58 & 68.57 \\

MiniLM     
& 0.937 & 0.934 & 0.942 & 0.938
& 127.26 & 10.11 & 98.95 \\

ALBERT     
& \textbf{0.944} & \textbf{0.943} & \textbf{0.945} & \textbf{0.944}
& \textbf{44.58} & 32.18 & 31.08 \\
\bottomrule
\end{tabular}
}
\end{table}

\begin{figure}[t]
\centering
\includegraphics[width=\columnwidth]{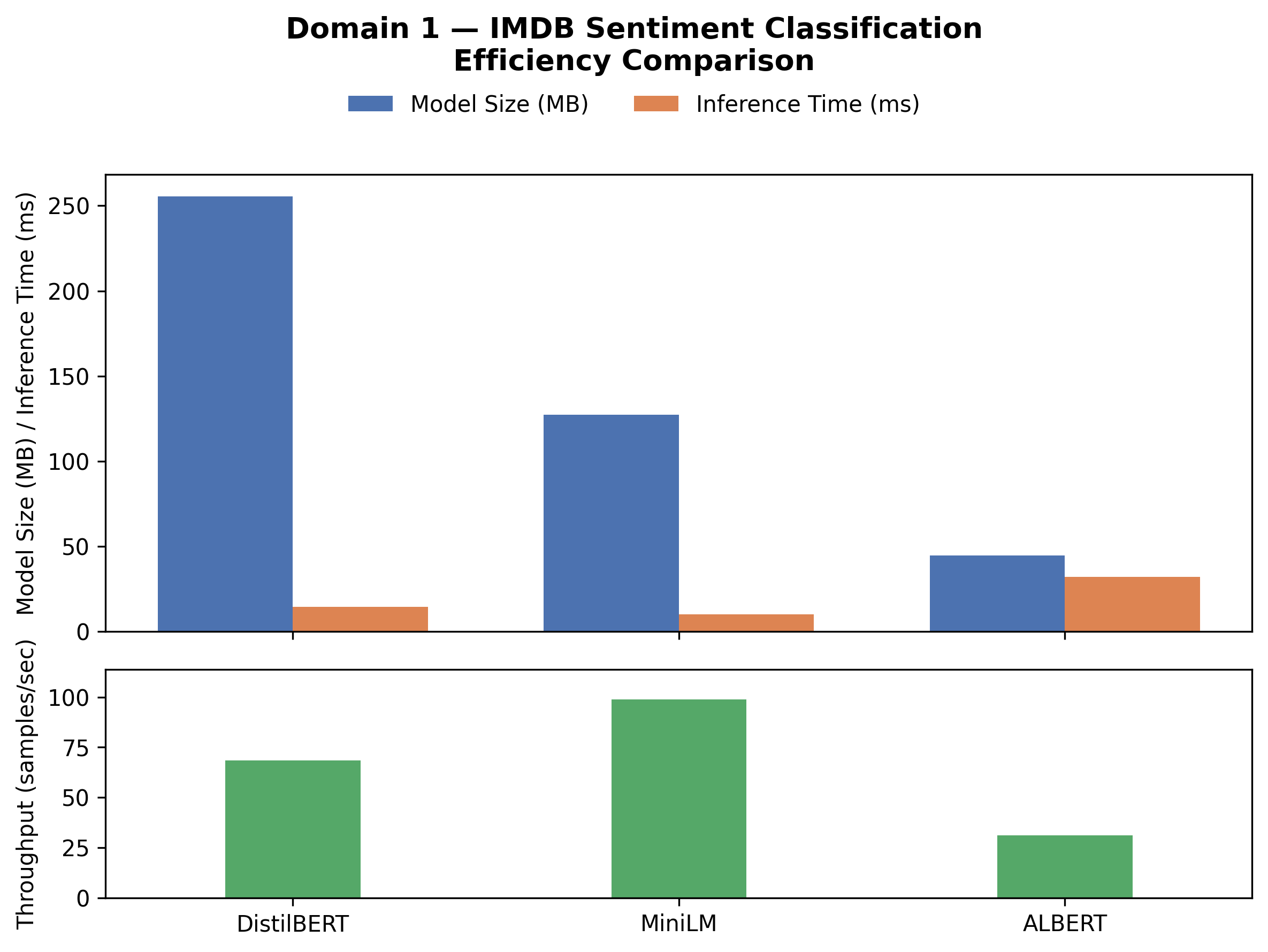}
\caption{Comparison of inference latency, throughput, and model size across models for Domain 1 (IMDB Sentiment).}
\label{fig:domain1}
\end{figure}

Confusion matrices for this domain are provided in Appendix~A.

\subsection{Domain 2: News Topic Classification (AG News)}

For the four-class AG News topic classification task, ALBERT achieves the highest classification accuracy, with DistilBERT and MiniLM closely following, indicating that topic categorization primarily relies on coarse semantic cues rather than fine-grained contextual reasoning. MiniLM delivers the highest throughput, supporting rapid large-scale document categorization in news aggregation and content routing systems where latency is critical.

\begin{table}[t]
\centering
\caption{Performance and Efficiency Metrics — Domain 2 (AG News)}
\label{tab:domain2}
\resizebox{\columnwidth}{!}{
\begin{tabular}{lccccccc}
\toprule
\textbf{Model} 
& \multicolumn{4}{c}{\textbf{Accuracy Metrics}} 
& \multicolumn{3}{c}{\textbf{Efficiency Metrics}} \\
\cmidrule(lr){2-5} \cmidrule(lr){6-8}
& Acc. & Prec. & Rec. & F1 
& Size (MB) & Inf. (ms) & Thpt. \\
\midrule
DistilBERT 
& 0.946 & 0.946 & 0.946 & 0.946
& 255.42 & 3.30 & 302.81 \\

MiniLM     
& 0.947 & 0.948 & 0.947 & 0.947
& 127.26 & \textbf{2.14} & \textbf{466.88} \\

ALBERT     
& \textbf{0.949} & \textbf{0.949} & \textbf{0.949} & \textbf{0.949}
& \textbf{44.58} & 7.48 & 133.66 \\

\bottomrule
\end{tabular}
}
\end{table}

\begin{figure}[t]
\centering
\includegraphics[width=\columnwidth]{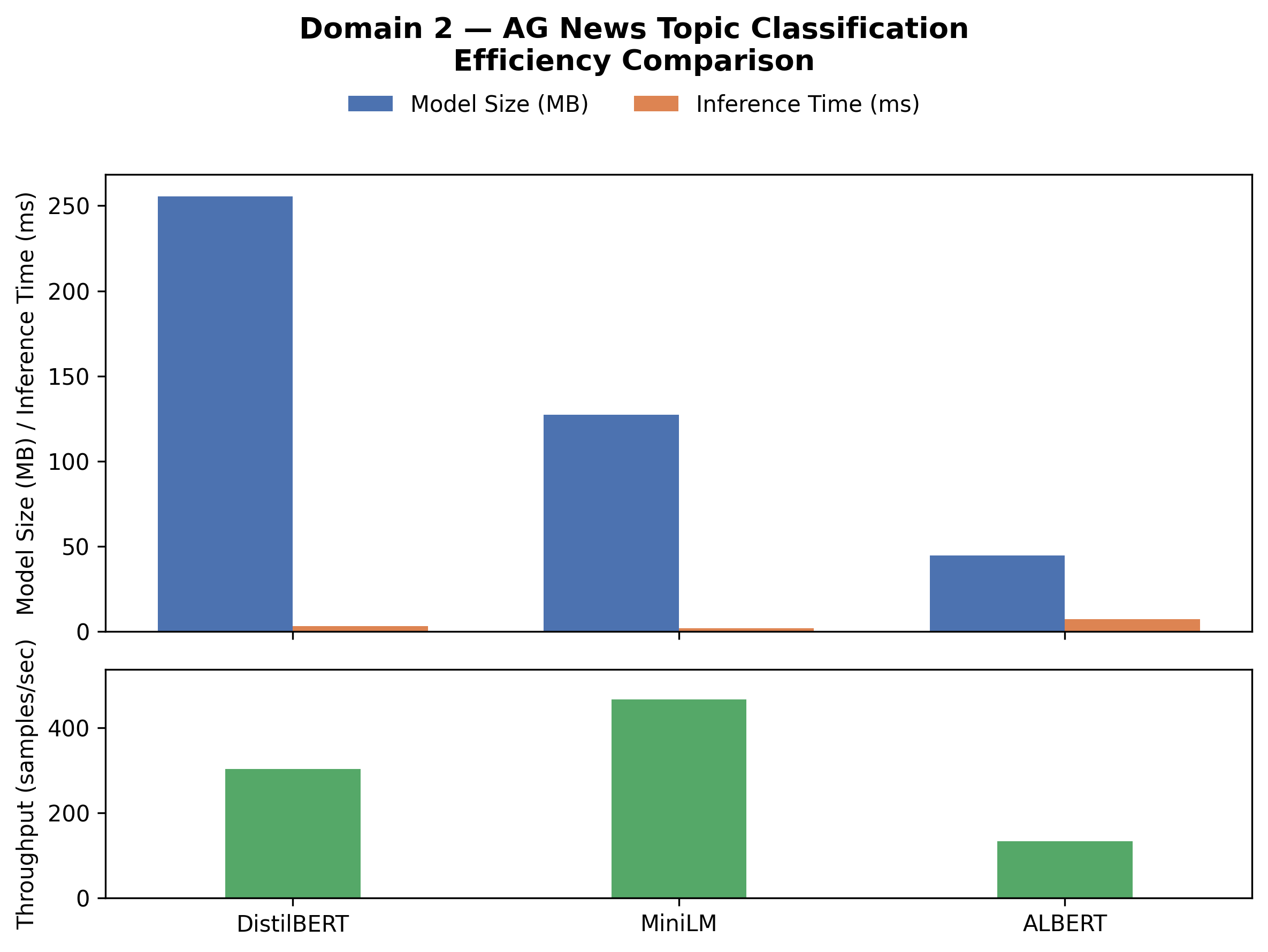}
\caption{Comparison of inference latency, throughput, and model size across models for Domain 2 (AG News).}
\label{fig:domain2}
\end{figure}

Confusion matrices for this domain are provided in Appendix~B.

\subsection{Domain 3: Toxicity and Hate Speech Detection}

The toxicity and hate speech detection task presents a more challenging, subjective classification problem. These thresholds are adopted as a pragmatic operational approximation commonly used in applied content moderation settings, enabling categorical evaluation while acknowledging that hate speech exists on a continuum.We do not claim these boundaries to be normative, and alternative threshold choices may affect class distributions and absolute performance metrics.
ALBERT achieves the highest overall accuracy and F1-score, reflecting strong generalization under label ambiguity despite its compact architecture. MiniLM provides the fastest inference and highest throughput, making it particularly suitable for real-time content moderation pipelines. ALBERT maintains competitive accuracy while offering the smallest model footprint, supporting deployment in low-resource or edge environments. Reduced F1-scores across all models reflect inherent annotation noise and class overlap within the dataset.

\begin{table}[t]
\centering
\caption{Performance and Efficiency Metrics — Domain 3 (Cyberbullying)}
\label{tab:domain3}
\resizebox{\columnwidth}{!}{
\begin{tabular}{lccccccc}
\toprule
\textbf{Model} 
& \multicolumn{4}{c}{\textbf{Accuracy Metrics}} 
& \multicolumn{3}{c}{\textbf{Efficiency Metrics}} \\
\cmidrule(lr){2-5} \cmidrule(lr){6-8}
& Acc. & Prec. & Rec. & F1 
& Size (MB) & Inf. (ms) & Thpt. \\
\midrule
DistilBERT 
& 0.949 & 0.948 & 0.949 & 0.949
& 255.42 & 2.71 & 368.75 \\
MiniLM     
& 0.906 & 0.897 & 0.906 & 0.900
& 127.26 & \textbf{1.85} & \textbf{540.17} \\
ALBERT     
& \textbf{0.951} & \textbf{0.950} & \textbf{0.951} & \textbf{0.950}
& \textbf{44.58} & 6.15 & 162.51 \\
\bottomrule
\end{tabular}
}
\end{table}

\begin{figure}[t]
\centering
\includegraphics[width=\columnwidth]{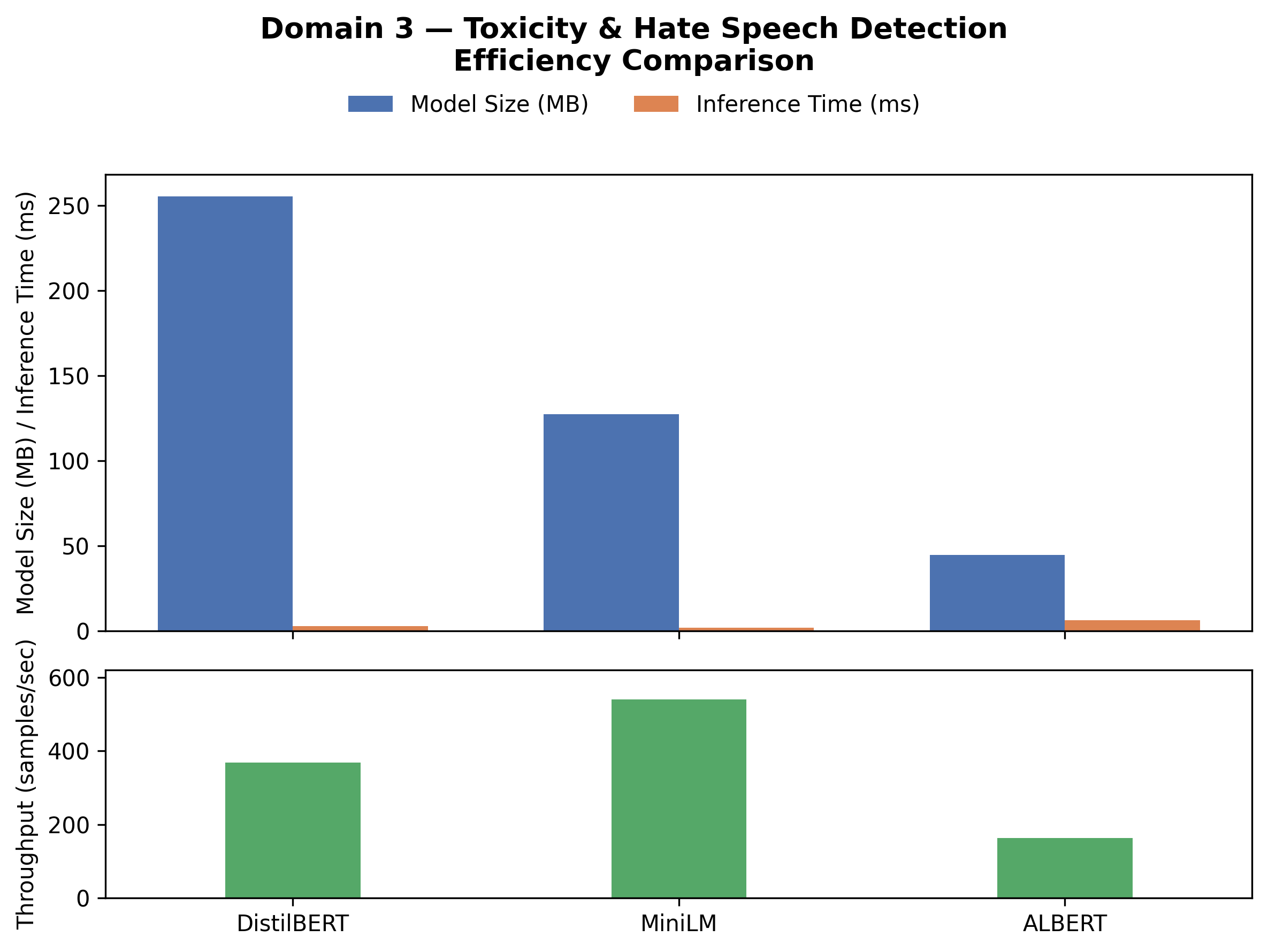}
\caption{Comparison of inference latency, throughput, and model size across models for Domain 3 (Cyberbullying).}
\label{fig:domain3}
\end{figure}

Confusion matrices for this domain are provided in Appendix~C.

\section{Cross-Domain Analysis}
Across all three evaluated domains—sentiment classification, news topic classification, and toxicity detection—consistent performance patterns emerge among the evaluated lightweight Transformer models. Across all three evaluated domains, no single model consistently dominates accuracy. ALBERT achieves the highest task-specific accuracy in multiple domains, while DistilBERT demonstrates stable and competitive performance across all tasks, indicating robust generalization despite reduced model depth. This consistency suggests that knowledge distillation effectively preserves task-relevant linguistic features across diverse text distributions.

MiniLM consistently delivers superior inference speed and throughput, often achieving 1.5--2$\times$ higher processing rates than competing models. This performance advantage remains stable across domains of varying complexity, reflecting the effectiveness of deep self-attention distillation in reducing computational overhead while maintaining competitive accuracy. Such consistency highlights MiniLM’s suitability for high-volume, latency-sensitive enterprise workflows.

ALBERT maintains the smallest model footprint across all tasks, demonstrating that parameter sharing and embedding factorization substantially reduce memory requirements. However, its comparatively slower inference speed reflects architectural trade-offs introduced by cross-layer parameter sharing, which increases sequential dependencies during forward passes \cite{lan2019albert}. Despite this limitation, ALBERT’s accuracy remains competitive, particularly in tasks dominated by coarse semantic cues.

Table~\ref{tab:crossdomain} summarizes the overall cross-domain comparison.

\begin{table}[t]
\centering
\caption{Overall Cross-Domain Comparison Summary}
\label{tab:crossdomain}
\begin{tabular}{lll}
\hline
\textbf{Category} & \textbf{Best Model} & \textbf{Rationale} \\
\hline
Accuracy (Task-Specific) & ALBERT & Highest accuracy in multiple domains \\
Consistency & DistilBERT & Stable performance across all tasks \\
Inference Speed & MiniLM & Lowest latency and fastest inference \\
Throughput & MiniLM & Highest samples processed per second \\
Model Size & ALBERT & Smallest memory footprint \\
Resource Efficiency & ALBERT & Minimal parameters with stable accuracy \\
\hline
\end{tabular}

\end{table}

These findings illustrate a consistent accuracy--efficiency trade-off: DistilBERT is best suited for precision-critical tasks, MiniLM for high-throughput deployment scenarios, and ALBERT for environments constrained by memory or storage limitations.

\section{Discussion}

While IMDB and AG News are standard academic benchmarks, they approximate enterprise workloads in sentiment monitoring, document routing, and automated content tagging. These datasets provide controlled, reproducible proxies for real-world text streams encountered in customer feedback systems, internal knowledge bases, and content moderation pipelines. These datasets are not claimed to be proprietary enterprise data; rather, they serve as standardized, reproducible proxies for enterprise NLP workloads such as sentiment monitoring, document routing, and content moderation.

The observed performance differences can be directly attributed to architectural design choices. DistilBERT’s strong accuracy arises from knowledge distillation, which transfers rich contextual representations from BERT into a shallower model while preserving semantic fidelity \cite{sanh2019distilbert}. This allows DistilBERT to generalize effectively across heterogeneous domains without excessive computational overhead.

MiniLM’s efficiency advantage stems from deep self-attention distillation, which prioritizes learning attention relationships rather than full hidden-state representations \cite{wang2020minilm}. This design choice significantly reduces hidden dimensions and computational cost, enabling consistently high throughput with only modest accuracy degradation. As a result, MiniLM offers a favorable balance for real-time and large-scale enterprise NLP systems.

ALBERT achieves its compactness through aggressive parameter sharing and factorized embeddings, drastically reducing model size \cite{lan2019albert}. While this design minimizes memory usage, it introduces sequential computation constraints that negatively impact inference latency. Nevertheless, ALBERT’s competitive accuracy suggests that such compression techniques remain viable for low-resource or edge deployments where memory constraints outweigh latency considerations.

Compared to larger contemporary models such as LLaMA or Mistral, these lightweight architectures offer substantial reductions in inference time, memory consumption, and operational cost. For example, MiniLM’s throughput enables processing approximately 1.5--2$\times$ more documents per second in enterprise pipelines, translating directly into reduced infrastructure costs and improved scalability \cite{zhang2024lightweight}.

\section{Limitations}

This study is subject to several limitations. First, all experiments are conducted on English-language datasets; performance may vary in multilingual or low-resource language settings. Second, evaluations are performed on a single GPU configuration, and results may differ under CPU-only deployment or distributed inference environments. Third, hyperparameter tuning is intentionally limited to reflect enterprise constraints, potentially leaving performance gains unexplored.

Additionally, results are reported from single-run evaluations without confidence intervals or statistical significance testing. While this aligns with deployment-oriented benchmarking—where throughput and latency are prioritized over repeated stochastic training—it limits formal statistical comparison.We do not claim statistical superiority between closely spaced accuracy values. Observed differences should be interpreted as indicative trends under fixed training conditions rather than statistically significant margins. In enterprise deployment contexts, such trends are often sufficient to inform architectural selection.Future work may address these limitations through multilingual evaluation, broader hardware profiling, and controlled variance analysis.

\paragraph{Reproducibility.}
All experiments were conducted using publicly available datasets and pretrained models. The complete source code, including training scripts, evaluation pipelines, and experimental configurations, is publicly available at  \url{https://github.com/shahmeer07/enterprise-nlp-lightweight-transformer-benchmark}.

\section{Conclusion}

This work presents a deployment-oriented benchmark of lightweight Transformer models under rapid fine-tuning constraints. Through evaluation across sentiment analysis, news classification, and toxicity detection tasks, We demonstrate that no single model dominates all dimensions of performance. DistilBERT offers the most consistent accuracy, MiniLM delivers superior inference efficiency, and ALBERT provides the smallest model footprint.

Rather than advocating a single best model, this study provides a practical decision framework for enterprise NLP deployment, enabling practitioners to select models based on accuracy requirements, latency constraints, and resource availability. By integrating accuracy and efficiency metrics across multiple domains, this benchmark offers actionable guidance for deploying scalable NLP solutions in real-world enterprise systems.

\bibliographystyle{unsrt}  
\bibliography{references}

\appendix
\section{Domain 1 Confusion Matrices}
This appendix presents the confusion matrices for the customer sentiment classification task (IMDB dataset) using DistilBERT, MiniLM, and ALBERT. These matrices provide insight into class-wise prediction behavior and misclassification patterns.

\begin{figure}[h!]
\centering
\includegraphics[width=0.95\textwidth]{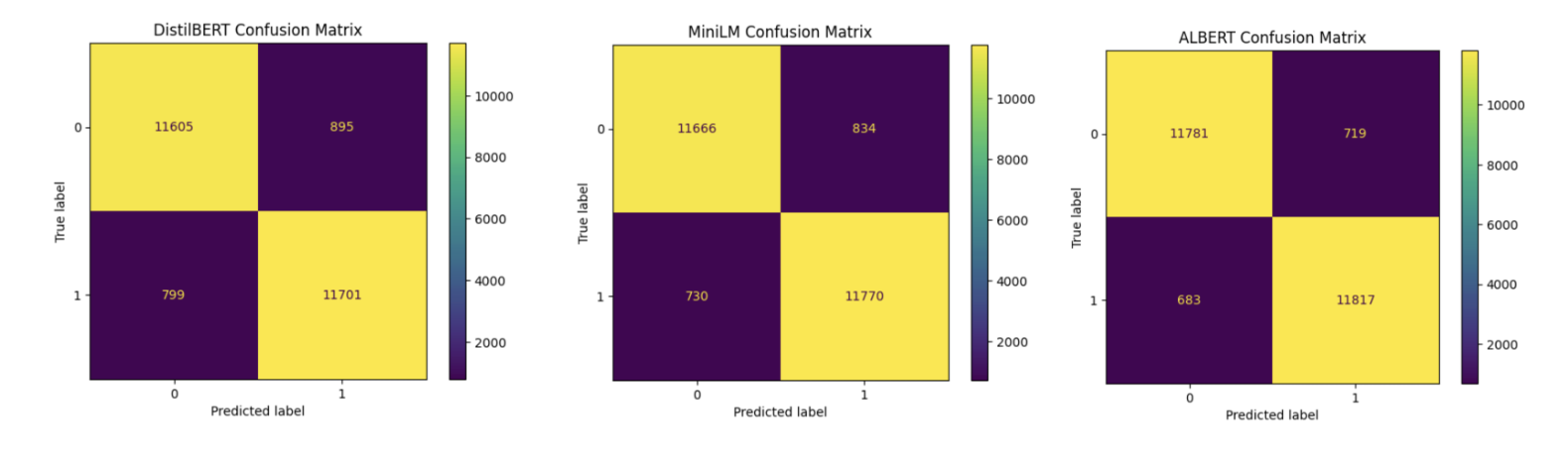}
\caption{Confusion matrices for Domain 1 (IMDB Sentiment Classification) using DistilBERT, MiniLM, and ALBERT.}
\label{app:domain1_cm}
\end{figure}

\section{Domain 2 Confusion Matrices}
This appendix reports confusion matrices for the four-class news topic classification task using the AG News dataset. The results highlight inter-class confusion across World, Sports, Business, and Sci/Tech categories.

\begin{figure}[h!]
\centering
\includegraphics[width=0.95\textwidth]{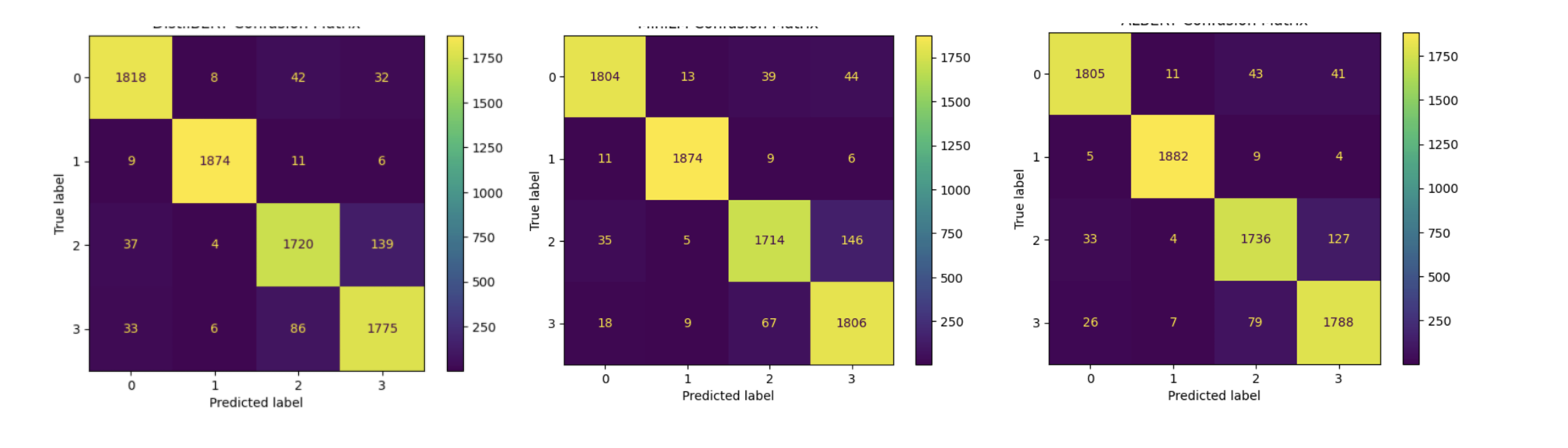}
\caption{Confusion matrices for Domain 2 (AG News Topic Classification) using DistilBERT, MiniLM, and ALBERT.}
\label{app:domain2_cm}
\end{figure}

\section{Domain 3 Confusion Matrices}
This appendix presents confusion matrices for the three-class toxicity and hate speech detection task derived from the Measuring Hate Speech corpus. The matrices illustrate the impact of label ambiguity and class overlap in real-world moderation datasets.

\begin{figure}[h!]
\centering
\includegraphics[width=0.95\textwidth]{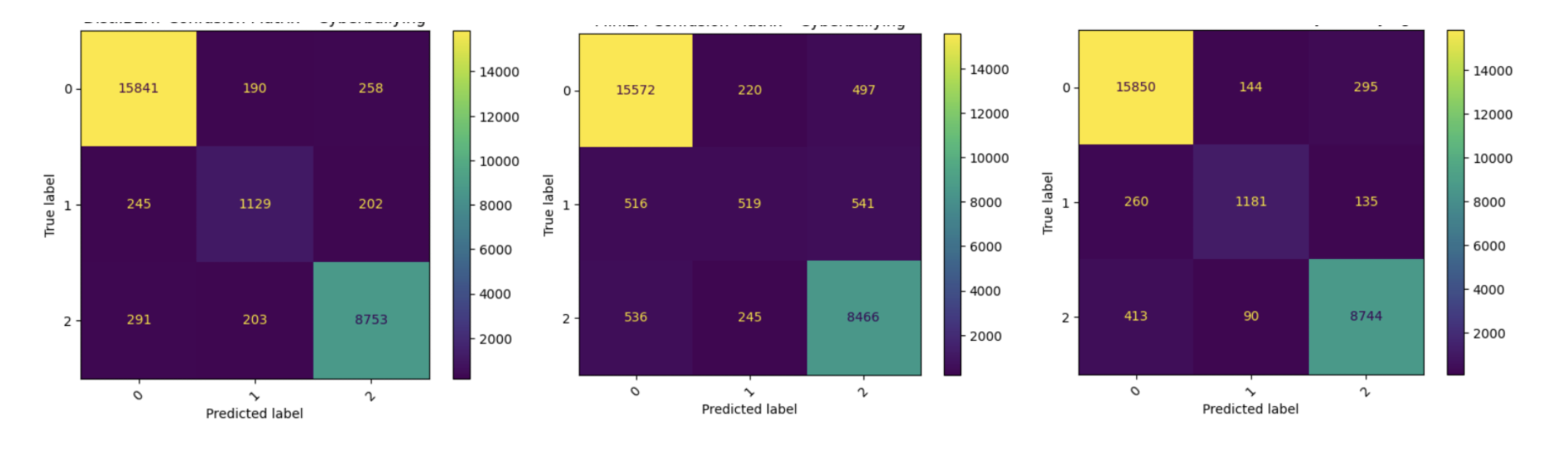}
\caption{Confusion matrices for Domain 3 (Toxicity and Hate Speech Detection) using DistilBERT, MiniLM, and ALBERT.}
\label{app:domain3_cm}
\end{figure}

\end{document}